\documentclass[siggraph,10pt]{acmart}
\renewcommand\footnotetextcopyrightpermission[1]{} 
\startPage{1}
 
\setcopyright{none}

\usepackage{booktabs}
\usepackage{subcaption}
\usepackage{listings}

\lstdefinestyle{mystyle}{
    basicstyle=\ttfamily\small,
    breakatwhitespace=false,         
    breaklines=true,                 
    captionpos=b,                    
    keepspaces=false,                 
    numbersep=2pt,                  
    showspaces=false,                
    showstringspaces=false,
    showtabs=false,                  
    tabsize=2
}
\begin{document}

\title[Short Title]{Automating Generation of Low Precision Deep Learning Operators}

\author{Meghan Cowan}
\affiliation{
  \department{University of Washington}
}
\email{cowanmeg@cs.uw.edu} 

\author{Thierry Moreau}
\affiliation{
  \department{University of Washington}
}
\email{moreau@cs.uw.edu} 

\author{Tianqi Chen}
\affiliation{
  \department{University of Washington}
}
\email{tqchen@cs.uw.edu} 

\author{Luis Ceze}
\affiliation{
  \department{University of Washington}
}
\email{luisceze@cs.uw.edu}

\begin{abstract}

State of the art deep learning models have made steady progress in the fields of computer vision and natural language processing, at the expense of growing model sizes and computational complexity. 
Deploying these models on low power and mobile devices poses a challenge due to their limited compute capabilities and strict energy budgets. 
One solution that has generated significant research interest is deploying highly quantized models that operate on low precision inputs and weights less than eight bits, trading off accuracy for performance. These models have a significantly reduced memory footprint (up to 32x reduction) and can replace multiply-accumulates with bitwise operations during compute intensive convolution and fully connected layers.

Most deep learning frameworks rely on highly engineered linear algebra libraries such as ATLAS or Intel's MKL to implement efficient deep learning operators.
To date, none of the popular deep learning directly support low precision operators, partly due to a lack of optimized low precision libraries.
In this paper we introduce a work flow to quickly generate high performance low precision deep learning operators for arbitrary precision that target multiple CPU architectures and include optimizations such as memory tiling and vectorization.
We present an extensive case study on low power ARM Cortex-A53 CPU, and show how we can generate 1-bit, 2-bit convolutions with speedups up to 16x over an optimized 16-bit integer baseline and 2.3x better than handwritten implementations.

\end{abstract}


\maketitle

\section{Introduction}
Over the past decade Deep Neural Networks (DNNs) have made substantial improvements in the fields of computer vision and natural language processing, with DNNs now surpassing human's in image recognition on the ImageNet dataset\cite{he2015delving}.
As the accuracy of state of the art DNNs has improved, model sizes and compute complexity has also increased. 
For example, AlexNet\cite{krizhevsky2012imagenet} uses roughly 60 million parameters and Facebook's DeepFace\cite{parkhi2015deep} uses 120 million, requiring billions of operations.
For low powered devices such as phones and edge devices, running inference on state of the art models poses a significant challenge due to their compute and memory requirements. Large models must be stored in off-chip DRAM where fetching model parameters from memory can become the dominant energy and time consumer\cite{han2016eie}, and computation is dominated by expensive multiply-accumulate operations.

To combat this, prior work has explored trading accuracy for performance by using low precision weights and activations of a few bits, or even a single bit as is the case with Binarized Neural Networks (BNNs)\cite{rastegari2016xnor}\cite{courbariaux2016binarized}.
Low precision models are drastically smaller, greatly reducing the cost of fetching model parameters from memory. For example BNN models are up to 32x smaller than full precision models and can often fit in on chip memory \cite{rastegari2016xnor}.
During compute intensive dense and convolution layers, multiplication can be replaced with cheap bitwise operations and popcount by computing in Hamming space, allowing for high performance inference. 

While training low precision networks has received lots of interest in the research community and made steady progress towards improving accuracy\cite{zhou2016dorefa}\cite{cai2017deep}, most work assumes low precision inference will result in a linear speedup over full precision and reference theoretical speedups based off of number of operations. However, achievable speedups are dependent on many factors including architectural support for matrix operations and software optimizations to effectively use the underlying memory subsystem and hardware instructions. 
Current deep learning frameworks leverage decades of prior work in optimizing linear algebra operations through libraries such as NNPACK\cite{nnpack} and Intel's MKL \cite{intel-alt} that optimize for both hardware backends and matrix properties, resulting in highly efficient floating point and integer operations. Naive implementations of bitserial operators can be slower than 8-bit integer and even full precision floating point implementations, partly because they lack optimized operator implementations.

In this paper we address the challenges of writing optimized low precision bitserial operators for CPUs.
We introduce a work flow to quickly generate high performance low precision deep learning operators for arbitrary precision that target multiple CPU architectures and include optimizations such as memory tiling and vectorization.
Specifically our contributions include:
\begin{itemize}
    \item A library of operators for quantization of floating point data, flexible \emph{bit packing}, \emph{bitserial} matrix multiply, and convolution.
    \item An extensive case study of optimizing low precision operators for a low power Raspberry Pi 3B that can surpass state of the art hand written 16-bit kernels for 1-bit weight, 2-bit activation convolutions.
\end{itemize}

\section{Background and Related Work}

\subsection{Low Precision Neural Networks}
Low precision neural networks operate on weights and activations quantized down to a few bits, or a single bit in the extreme case of BNNs.
These types of neural networks improve performance over full precision models by significantly reducing memory movement costs and computing using \emph{bitserial} methods and can be deployed on existing hardware.

BNNs, such as XNORNet\cite{rastegari2016xnor} and BinaryNet\cite{courbariaux2016binarized} have achieved near state of the art results on datasets such as MNIST and CIFAR-10, showing that binarization works extremely well for relatively simple datasets. However, they preform significantly worse than full precision models for complex datasets like ImageNet, with accuracy degradations over 18\% for XNORNet on ResNet18\cite{he2016deep}.
To combat this, recent work\cite{zhou2016dorefa}\cite{hubara2016quantized} has shifted to low precision networks which relax activation quantization to a few bits, while keeping weights binarized. 
Current state of the art low precision networks have made significant improvements in the accuracy. HWGQ\cite{cai2017deep} have shown that 1-bit weights and 2-bit activations models can achieve top-1 accuracy drops of between 5 to 9\% on different models for ImageNet, including ResNet18 and GoogLeNet\cite{fromm2018heterogeneous}\cite{szegedy2015going}.

\subsection{Bitserial Computation}
While FPGAs and custom hardware can take advantage of arbitrary precision through custom datapaths \cite{umuroglu2017finn}, most commodity hardware such as CPUs, have no support for low precision data types. Efficient handling of low precision data requires \emph{bit packing} into a larger storage data type such as an 8-bit or 32-bit integer, and \emph{bitserial} operations that compute on an implicit vector of packed quantized data.

We describe bitserial dotproducts as seen in previous work\cite{umuroglu2017towards}\cite{zhou2016dorefa}, starting first with binary case and extending towards higher precisions.
A binary dot product between two vectors containing only elements of 0 and 1, can be computed by bit-wise anding the two vectors and counting the number of 1's in the result using popcount, as seen in Equation \ref{Equation 1}. If binary data is encoded in the bipolar format (-1 and 1), then bitwise xnor replaces bitwise and.
\begin{align}
	\boldmath{x} \cdot \boldmath{y} = popcount(x \& y)\ 
    \label{Equation 1}
\end{align}

Binary dot products can be easily extended to preform \emph{bitserial} dot products between an M-bit and N-bit vector, by computing the weighted sum of MN binary dot-products as described in Equation 2, where n and m refer to the bit position of x and y.

\begin{align}
	 \boldmath{x} \cdot \boldmath{y} = \sum_{n=0}^{N-1} \sum_{m=0}^{M-1} popcount(x_n \& y_m) * 2^{m+n}
\label{Equation 2}
\end{align}

While bitserial dot products can be used for any precision inputs, its compute complexity grows linearly with the product of x and y's bitwdiths as O(MN), so this technique is only practical for very quantized data.
Equation \ref{Equation 2} can be modified to support signed data by weighting binary dotproducts by their sign.

On CPUs low bit datatypes are not supported and data must be \emph{bitpacked} to efficiently support bitserial computation and take full advantage of the hardware's data path.
This is achieved by splitting input vectors into individual bitplanes essentially creating B binary vectors. The binary vectors are then packed data into a larger storage type such as a \texttt{uint32}. Bitwise operations can be directly applied to the packed data, allowing low precision data to be efficiently stored and computed. 

\begin{figure}[!hbt]
\centering
\includegraphics[width=0.45\textwidth]{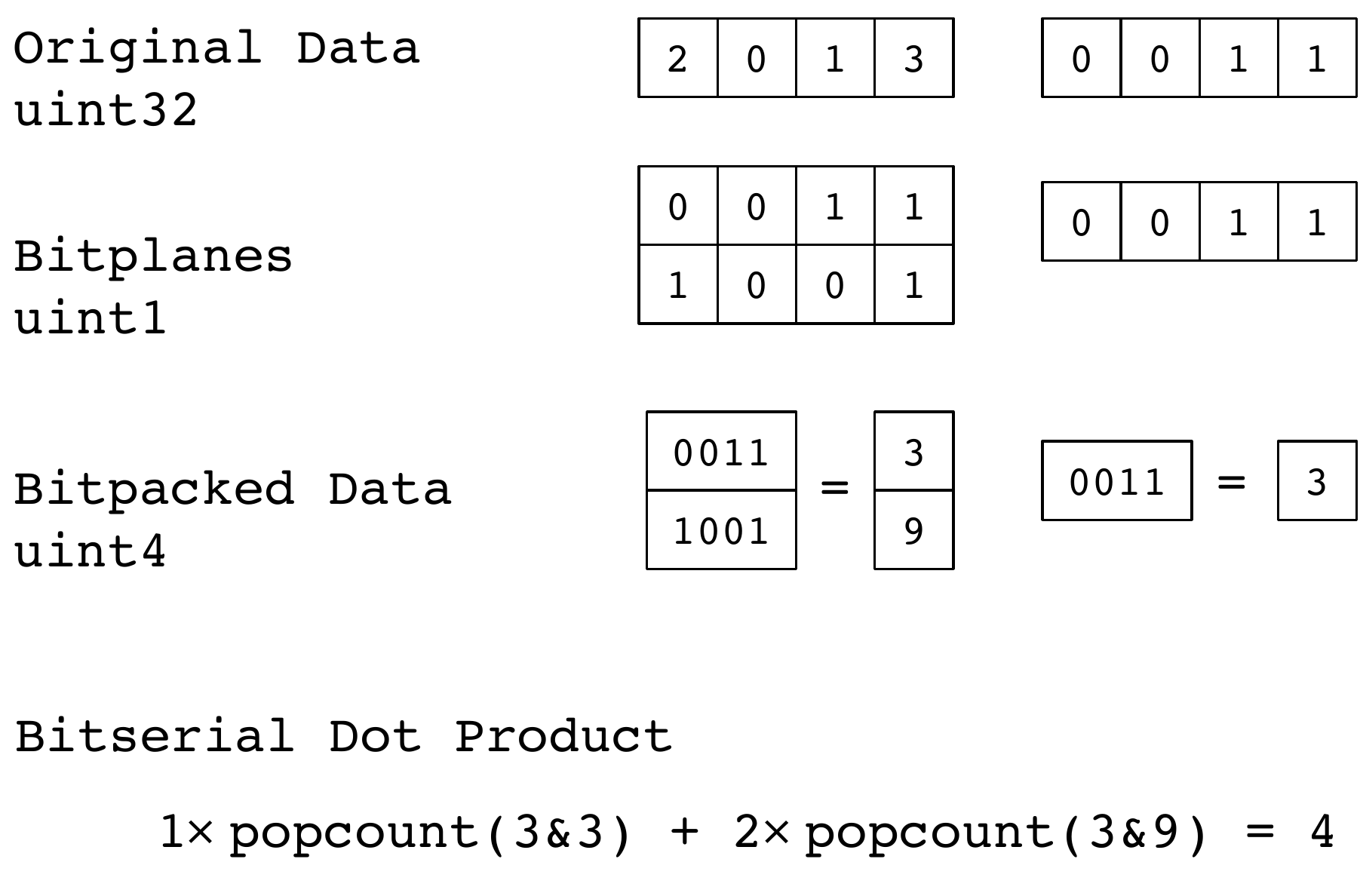}
\caption{Example of preforming a packed bitserial dot product. Input vectors are separated into bit planes which are then packed into a larger storage type. A weighted sum of binary dot products is then preformed between the each bit plane.}
\label{fig:simple-pack}
\end{figure}

In Figure \ref{fig:simple-pack} we show an example of how bitpacking and bitserial computation work on a 2-bit, 1-bit dot product.
Input data is first \emph{bitpacked} by separating the bits of each element into separate bitplanes. Each bitvector is then compressed into a larger integer type, for simplicity a four-bit unsinged integer. Since bitwise operations have no carry chains, it can be preformed on packed data in an implicit vectorized fashion and take full advantage of the hardware's data path. For example if data into a 32-bit integer, a single pair of popcount-and instructions computes the 32 element dot product.


Operators are specified in a declarative tensor expression language, that is separate from the schedule optimizations that implement it (a model based off of Halide\cite{ragan2013halide})


\section{Low Precision Operators}


\begin{figure}
\centering
\includegraphics[width=0.35\textwidth]{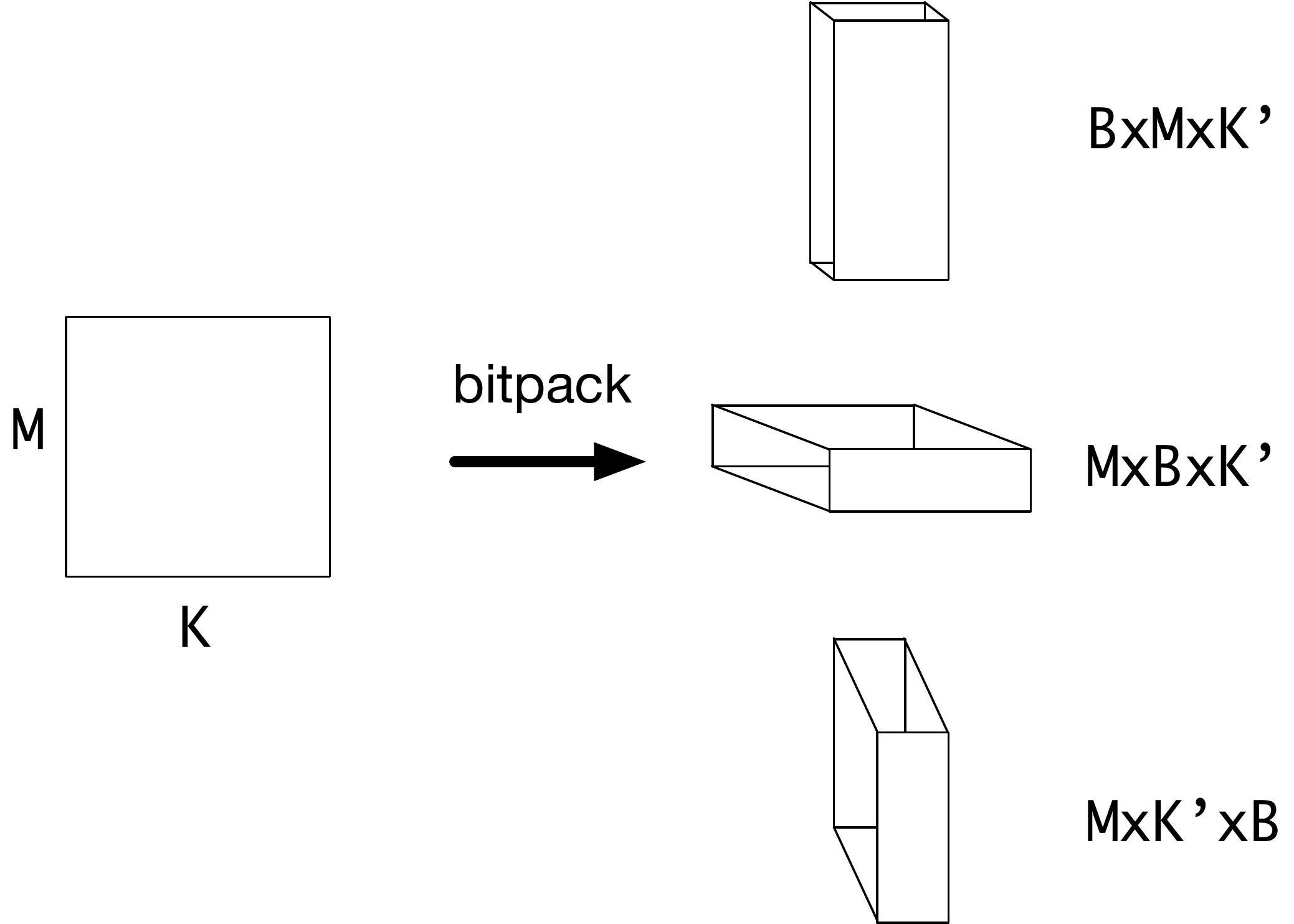}
\caption{Example of resulting packed tensors from bit packing a MxK matrix. Our bitpacking operator supports flexible placement of the bit axis dimension.}
\label{fig:bitpacking}
\end{figure}


Low precision operators rely on efficient bitserial computation. We implement our operators using TVM, the deep learning compiler\cite{chen2018tvm}. 
Our operators are designed to provide flexibility in precision and data layout, and performance portability across different CPU architectures. 
In Halide fashion, we provide (1) a declarative computation rule that describes the transformation of input to output tensors and (2) a separate schedule that contains the implementation of how to the data is transformed. We leverage TVM features to provide high-level architecture agnostic optimizations for parallelism and memory tiling. 
For CPU code generation, TVM uses LLVM, which we find produces adequate code. However, for further performance we implement architecture-specific mickrokernels that take advantage of unique hardware intrinsics that LLVM misses. 


\subsection{Preprocessing: quantization and bitpacking} 
In order to apply bitserial operators, both input activation and weight tensors need to be in the correct format which involves quantizing data to the desired precision and bitpacking them to the proper data layout.

Since bitserial operators compute on each bit of an element individually, the separate \emph{bitplanes}, the set of bits for each bit position, of each input tensor must easily accessible.
This is accomplished by a \emph{bitpacking} step that transforms the input tensors into a \emph{bitpacked tensor}. Elements of the packed tensor hold a single bit for many elements as opposed to all bits of a single element for the input tensor.


 We provide a bitpacking operator that takes an arbitrary $d$ dimension tensor and returns a $d+1$ dimension bitpacked tensor, with a new \emph{bit axis} addressing the bit-planes of the tensor. Though the bitpacked tensor has more dimensions its total size is smaller as multiple elements are packed along \emph{reduction axis} into a single element. The tensor's reduction axis is dependent on the layout and represents the axis along which elements are multiplied and accumulated against.

Our bitpacking operator is flexible with respect to input data layout and size. The user specifies the input tensor's reduction axis and the bit-axis location and datatype of the packed tensor.  For example, in Figure \ref{fig:bitpacking}, we show an example of bitpacking matrix A, a 2-bit $m \times k$ matrix, where $K$ is the reduction axis. The packed matrix can have layouts $MK'B$, $MBK'$, or $BMK'$. 

The different layouts provide different types of data locality. When implementing bitserial operators, we found that flexibility in specifying bit position to be useful, as different bitserial algorithms rely on a specific packed layout
For example, work in\cite{umuroglu2017towards} relied an interleaved layout with B as the inner-most dimension while work in\cite{tulloch2017high} required B to be the outermost dimension 

\subsection{Low precision operators and high level optimizations:} 
We implement a library of low precision operators for common neural network operations such as 2D convolutions and dense matrix multiply, for arbitrary CPU backends. These operators transform bitpacked tensor into standard format tensors in a higher machine supported precision.
Using TVM we can leverage existing features of the compiler to cleanly describe our operator's implementation and quickly add optimizations for memory access, parallelism, and more. 

At the compute level we describe variations of bitserial convolutions that accept different high level data layouts such as NCHW and NHWC, and various convolution implementations such as lowering to matrix multiply and an efficient in place convolution modified from \cite{jiang2018efficient}, all parameterized for different precisions. Depending on the input's shape and characteristics of the convolution kernel, different implementations preform better. 


For each operator variation, we provide a generic schedule (described in section 4) for each these operators that creates moderately efficient code for any CPU backed that takes advantage of vectorization, tiling, and other CPU agnostic optimizations. 
While many of these techniques are well established and simple in theory, they are time consuming to implement, with some requiring re-writes of the entire implementation and tuning parameters such as tile sizes.
Furthermore, many of these parameters do not affect performance independently and must be tuned together, leading to a large search space of parameters to find an optimal configuration. We take advantage of TVM's auto-tuning capabilities\cite{chen2018learning} to search for optimal or near optimal parameter configurations.

\subsection{Architecture specific optimizations:}
TVM relies on LLVM to preform code generation to the desired CPU backend. Since generating optimal code is a hard problem, unsurprisingly, LLVM tends to produce sub-optimal code. To approach or surpass the performance of the state of the art operator implementations, schedules can be augmented with handcrafted custom microkernels to implement the core computation. This allows us to take advantage of new hardware intrinsics many CPU vendors are releasing to accelerate deep learning such as pairwise addition instructions, as well as optimize low level memory loads and stores. 

We create a custom ARM schedule (described in section 5) that takes advantage of specific hardware intrinsics to implement code that out-preforms state of the art hand optimized libraries. The custom schedule relies on highly optimized hand-generated microkernel.

\section{CPU-Agnostic Schedule}
\begin{figure}
\centering
\includegraphics[width=0.45\textwidth]{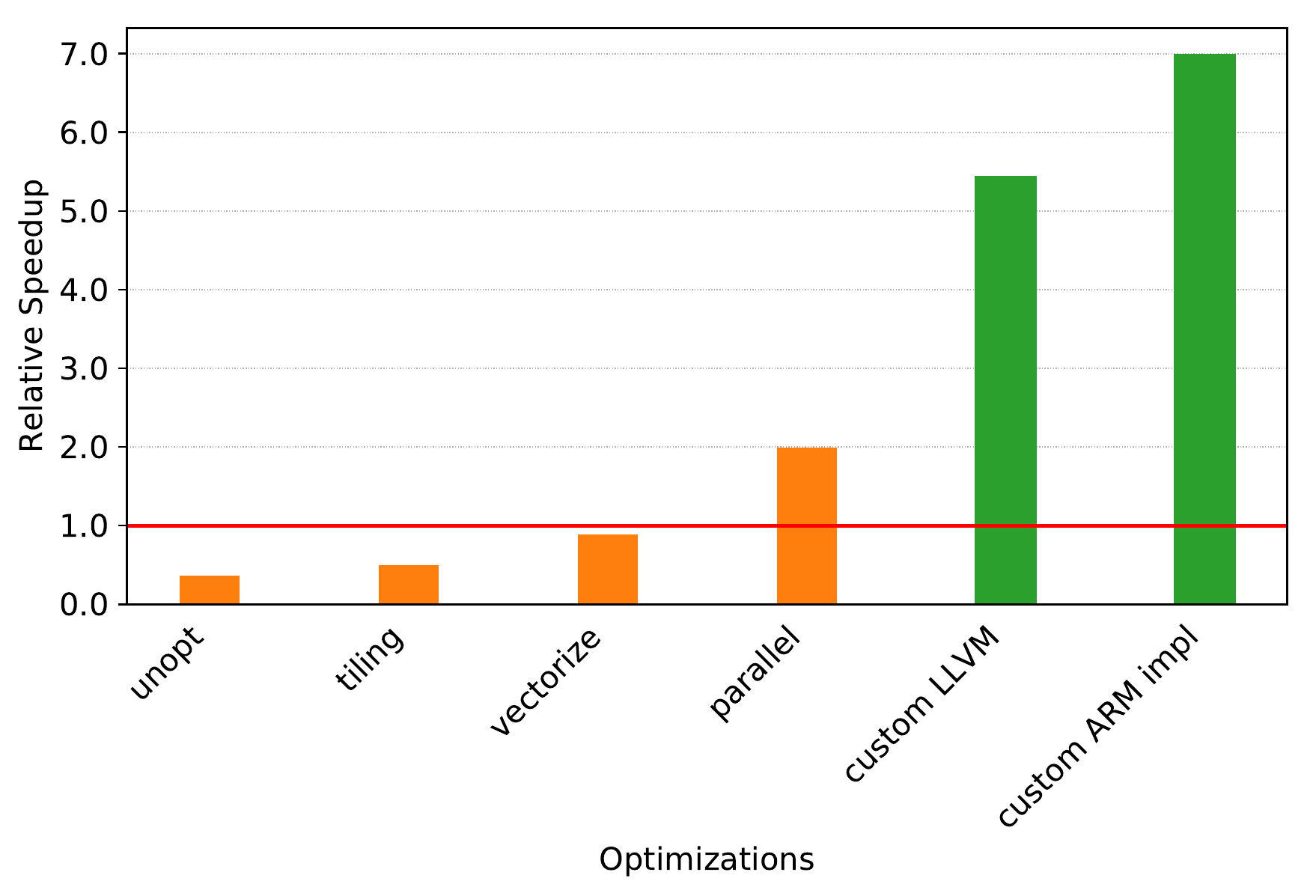}
\caption{Relative speedsup against 16-bit baseline for ResNet18 layer 2 on the Raspberry Pi, starting with an unoptimized schedule and adding optimizations incrementally. Green bars indicate optimizations part of the generic CPU schedule while green bars are ARM-specific optimizations.}
\label{fig:optimize}
\end{figure}

\begin{figure*}
\centering
\includegraphics[width=\textwidth]{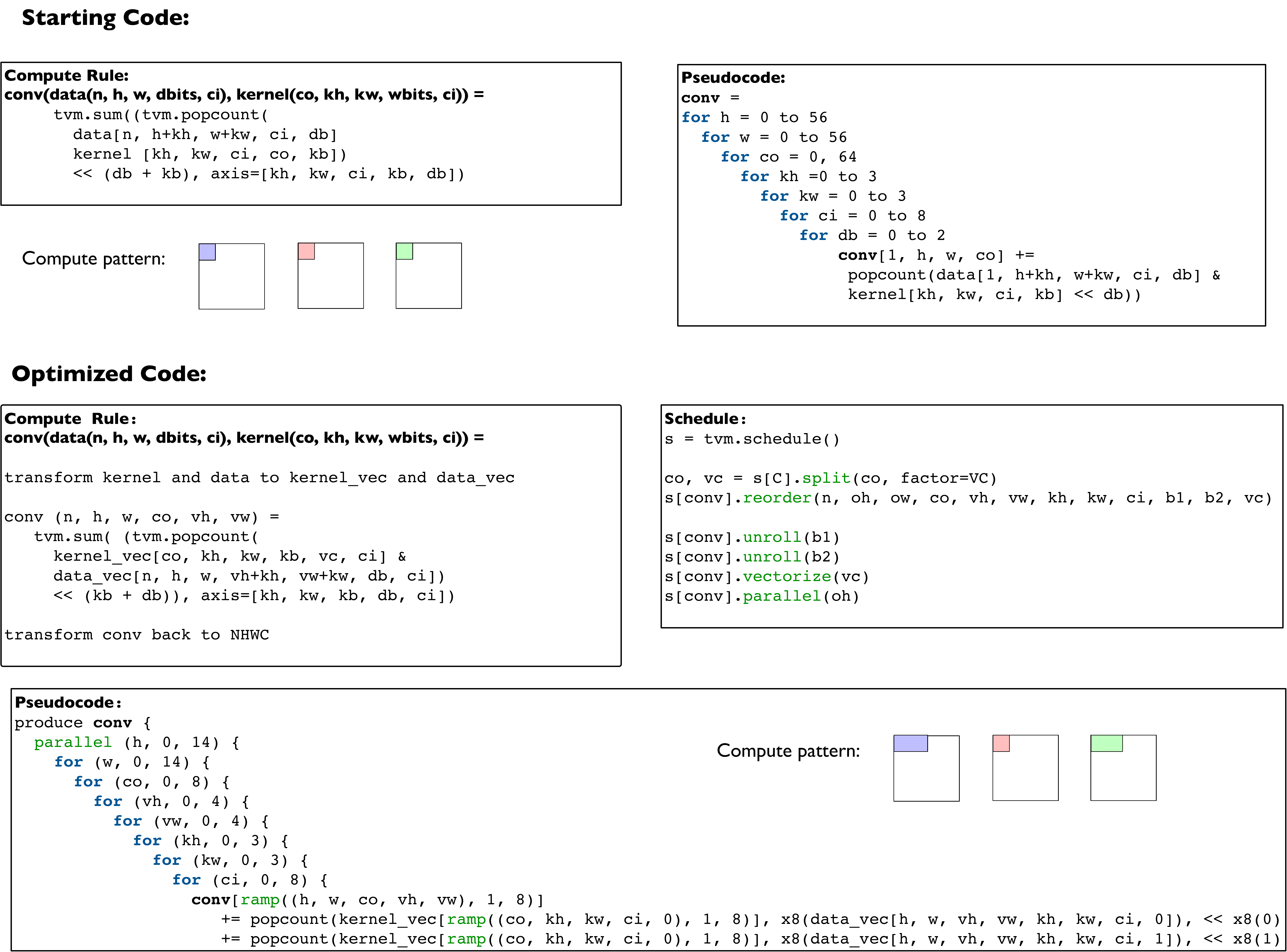}
\caption{On the right TVM pseudocode of relevant code to apply loop unrolling, vectorization, tiling, and parallelism, and the output C pseudocode on the left produced showing the affect of the schedule transformations.}
\label{fig:code}
\end{figure*}

In this section we describe the layers of optimizations that we explored when putting together our bitserial operator schedule template.
We show a running example of how we add in optimizations, and show their effects on a Raspberry Pi running the 2nd layer of ResNet18 (see Table 1 for details), for a 1-bit weight 2-bit activation convolution. Figure \ref{fig:code} shows snippets of the unoptimized compute rule and the pseudocode it generates, and the final optimized compute rule and schedule. In Figure 
\ref{fig:optimize} we show speedups against an optimized 16-bit integer convolution as we add in each optimization, showing how we can take an unoptimized schedule that is 0.36x slower than the baseline and make it almost 2x faster.
Many of these optimizations come from high performance computing techniques to improve memory locality and take advantage of parallelism and vectorization, but are necessary steps for achieving optimized low precision convolutions.

\paragraph{Tiling}
Tiling is a common optimization to improve temporal memory locality and reduce memory loads during matrix computations.
Tiling splits input tensors into multiple sub-tensors or tiles, such that each tile fits in the cache to avoid evicting elements that will be reused.

TVM exposes functions to easily split and reorder computational axis without the user needing to manually edit for loops. 
In Figure \ref{fig:code}a, we show how we apply tiling in TVM through calls to \texttt{split}, and \texttt{reorder}. The height, width, and channels of the output matrix are split into blocks of size VH by VW by VC elements, which can be tuned by TVM. Using reorder, the iteration order is updated so that computation is completed within each tiles.

After tiling, data is not fetched sequentially from input tensors as data is stored in row-order. A common trick to improve tiling's performance further, is to repack input tensors in a hierarchical fashion such that elements within a tile are stored sequentially, and the tiles then stored in a sequential order. Repacking requires a change to the computation description and is shown in the optimized compute rule in Figure \ref{fig:code}.

\paragraph{Unrolling}
Loop unrolling replaces loops with copies of the loop statement, and improves performance by reducing the number of branches during program execution. This performance gains comes at a cost of increased binary size as instructions inside the loop are duplicated. Loop unrolling is easily expressed in TVM through the \texttt{unroll} scheduling primitive, and we apply it small dimensions such as the bit axis. 

\paragraph{Vectorization}
Vectorization takes advantage of hardware support for SIMD instructions, and applies each instruction to a vector of data elements. Calling TVM's \texttt{vectorize} primitive on an axis provides a hint to the compiler to output vectorized instructions.
In our schedule, we tell the compiler to vectorize the innermost computation axis, since it is stored sequentially on output tensor and kernel tensor, allowing for contiguous vector loads and stores. 
In the generated C code in Figure \ref{fig:code}, single element indices are replaced with a \texttt{ramp(n)} expression representing a vector of n elements, or a vector made from n copies of a single element. 

On the ARM Cortex-A53, this allows us to use ARM's SIMD unit. The intrinsic popcount instruction, \texttt{vcnt}, is only available for SIMD instructions, so it is crucial for ARM devices to vectorize the binary dot product's computation or else popcount is inefficiently implemented in software. On X86, the opposite scenario occurs and popcount is only available in scalar registers. However, we found the vectorized software popcount outperforms the scalar hardware popcount as confirmed by \cite{mula2017faster}, and left vectorization in the generic CPU schedule.

\paragraph{Parallellization}
We parallelize the outer most axis of computation by calling TVM's \texttt{parallel} function, allowing us to take advantage of all four cores on the Raspberry Pi, so that the generic CPU schedule outperforms the baseline by almost 2x.

All of the optimizations described above (and a few more) are necessary to implement a high performance convolution implementation. Optimizing code is often time consuming to implement, and can significantly reduce the readability of code by through bloat from unrolling, and further nesting of loops from tiling. While we carefully decided the layout of data and which axes to tile, split, and parallelize, we rely on TVM to handle the time consuming process of generating the code, allowing us to quickly write moderately efficient schedules. 

\section{ARM Cortex A53 Specfic Convolution Schedule}
The techniques described in the previous section are CPU agnostic, with the same schedule able to generate code for different CPU backends such as X86 and ARM.
However, to reach the performance of hand optimized microkernels, schedules require architecture aware optimizations through back-end specific intrinsic instructions and human insight into efficient techniques. 

In this section we describe two ways in which we optimize ARM specific schedules. The first, a modification to the code generator that is invisible to the user and overrides LLVM's default popcount lowering rule. The second is an ARM specific \texttt{tensorize} primitive that implements a highly optimized bitserial matrix vector operation. 

\begin{figure}
\centering
\includegraphics[width=0.5\textwidth]{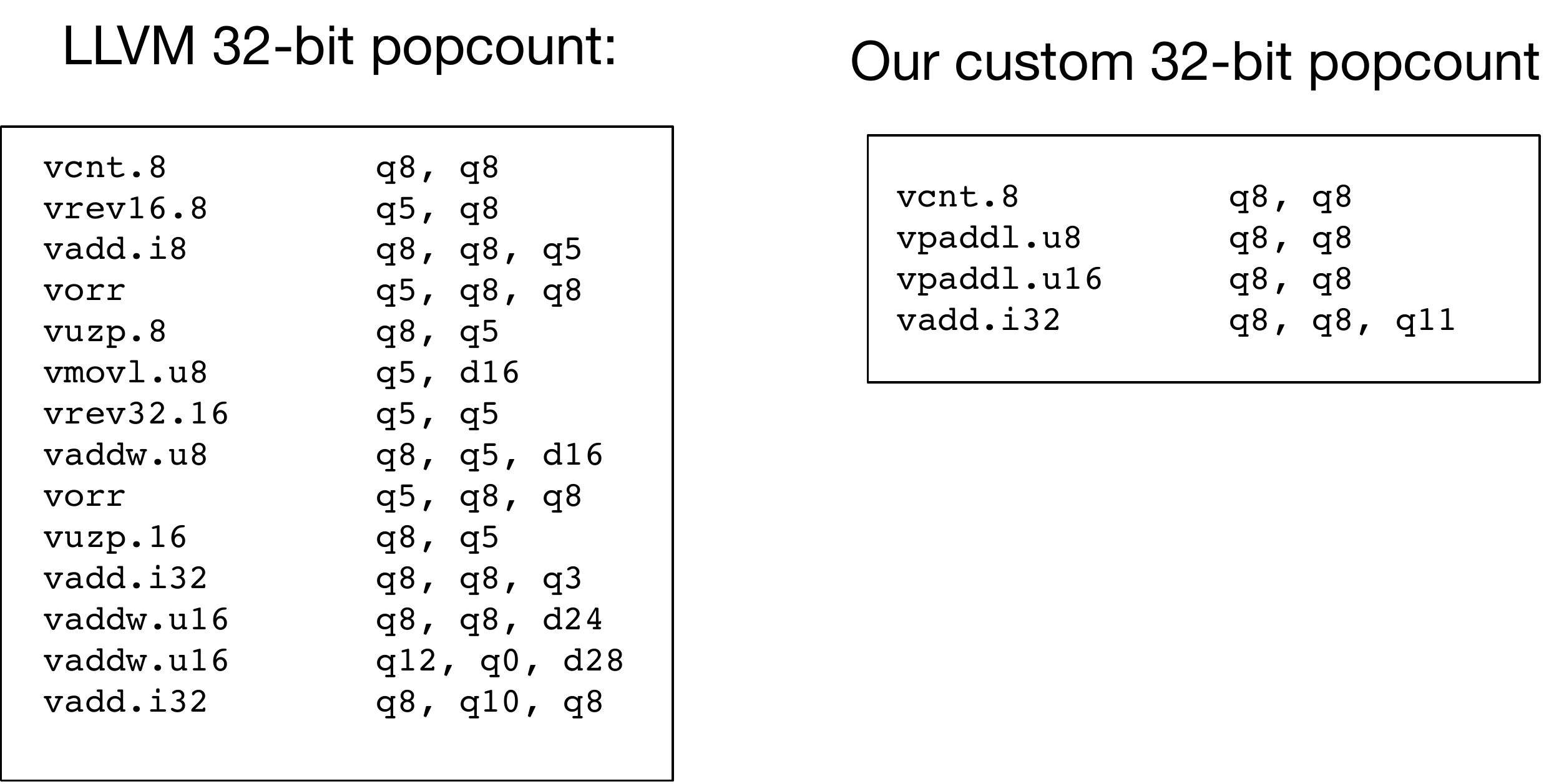}
\caption{Output ARM assembly code for a 32-bit vectorized popcount generated by LLVM vs. our custom lowering rule. Our results reduces the number of assembly instructions from 13 to 4.}
\label{fig:llvm}
\end{figure}

\begin{figure}
\centering
\includegraphics[width=0.5\textwidth]{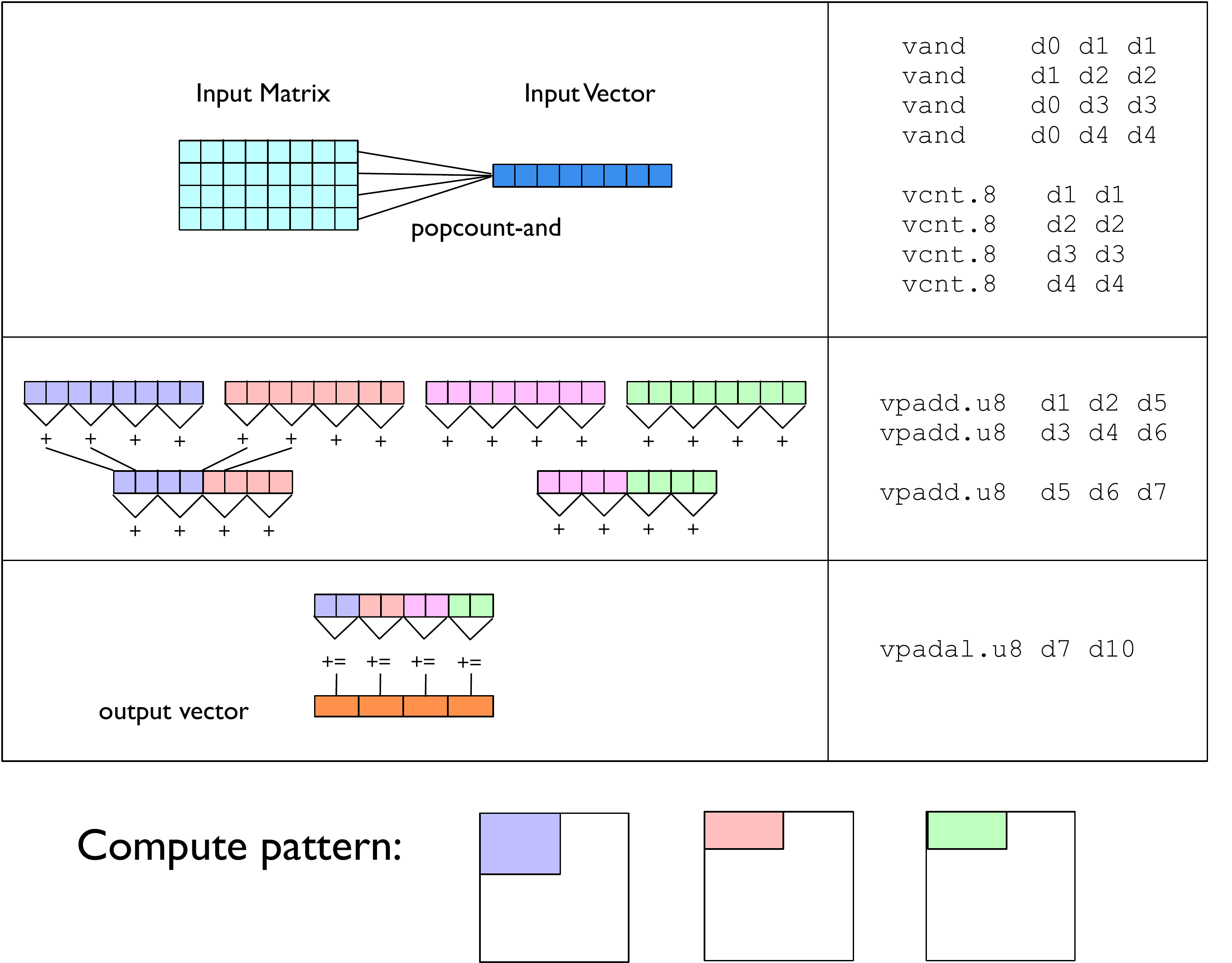}
\caption{Simplified example of our tensorized popcount reduction that uses and the specific ARM code it generates.}
\label{fig:tensorize}
\end{figure}

\subsection{Custom ARM Popcount Lowering Rule}
Analyzing generated assembly code revealed LLVM (TVM's code generator for ARM and other CPU backends, version 5.0) inefficiently lowers popcount, requiring 14 assembly instructions to implement a 32 bit vectorized popcount.
The inefficiency arises because ARM's hardware popcount instruction, \texttt{vcnt}, operates at a granularity of 8 bits. When LLVM lowers a 32-bit vectorized popcount, it must reinterprets the input as a vector of 8 bit, and then sum four neighboring elements to reform the 32-bit bector.
This default accumulation step is inefficient, requiring 13 additional assembly instructions. 

We replaced the default lowering rule in TVM's code generation for 32 bit vectorized popcounts with only 4 assembly instructions, by relying on a special intrinsic instruction, \texttt{vpaddl}, that performs pairwise adds in a larger datatype. Figure \ref{fig:llvm} shows the LLVM 32-bit popcount lowering rule as well as our custom rule. 

This change requires no modifications to the schedule, and significantly speeds up the generic CPU schedule by 2.7x on the Raspberry Pi, for a total speedup of 5.4x over the 16-bit integer baseline.

\subsection{Optimized ARM bitserial matrix vector inner loop}
In order to surpass the performance of hand-optimized code we analyzed code from Caffe2's ultra low precision library\cite{tulloch2017high} and gemmbitserial\cite{umuroglu2017towards}. We identified patterns TVM currently cannot express such as vectorization along a reduction access, and low-level tricks the code generator misses, and wrote an efficient bitserial matrix vector multiply micro-kernel.
We packaged the micro-kernel into \texttt{tensorize} primitive, which we parameterize for flexibility in activation/weight precision. Our tensorize primitive can be used in any ARM schedule, and we use it for all our ARM specific convolutions and matrix multiply schedules. However, it requires tiling input tensors such that to match vector lengths and relaying out inputs tensors such that tiles are stored contiguously.

Figure \ref{fig:tensorize} gives a simplified example of how our tensorized primitive behaves and the assembly code it generates. It preforms 4 bitserial dot products and optimizes accumulation and write back steps in a tree reduction fashion allowing for vectorized loads and stores on all tensors. 

We summarize the optimizations the code generator missed. We note that in high performance computing, it is common to in line assembly for performance critical sections, such as inner-loops of computation, and TVM provides a method to mimic this. 
\paragraph{Accumulation in small datatypes:} In order to prevent saturation of values the output of a low precision operation needs to be stored in a larger data type such as an \texttt{int16} or \texttt{int32} depending on the size of inputs tensors. 
TVM's code generator promotes inputs to a larger datatype after a single operation; however, multiple operations can be safely accumulated before moving up a storage size. 
Our tensorize primitive takes advantage of this by accumulating 8 bit integers until overflow is possible and then extending accumulation to 16 bit.

\paragraph{High level patterns:} Programmers are very good at writing code that follows high level patterns, such as the tree reduction patterns. In general compilers fail to preform high level optimizations such as this, since most optimizations occur between a small localized regions of code.

After adding the tensorized reduction to our inner-loop our low precision convolution schedule preforms almost 7x faster than the optimized 16-bit baseline, with the final 1.6x coming from ARM specific optimizations to the convolution schedule (see Figure \ref{fig:optimize}).

\section{Evaluation}

\subsection{Methodology}
We preform an analysis of our low precision bitserial operators on two CPU backends, a low-power Raspberry Pi with an ARM Cortex-A53 processor and a high-end x86 Intel i7-4790K processor.
The ARM Cotex-A53 is a four core 1.2 GHz  processor and belongs to the ARM-V8 architecture, while the X86 machine is a 4.0 GHz four core processor with eight hyperthreads. 

Note throughout the evaluation section when referring to our low precision operators we will adopt the naming convention from \cite{umuroglu2017towards} and refer to an x-bit weight and y-bit activation operation as WxAy. For all our experiments we include the cost of bitpacking activations with the quantized operations, but not the cost of bit-packing weights, since we assume weights can be bitpacked ahead of time. 

\subsection{Raspberry Pi Results}
\subsubsection{Matrix Multiply}
\begin{figure}
\centering
\includegraphics[width=0.5\textwidth]{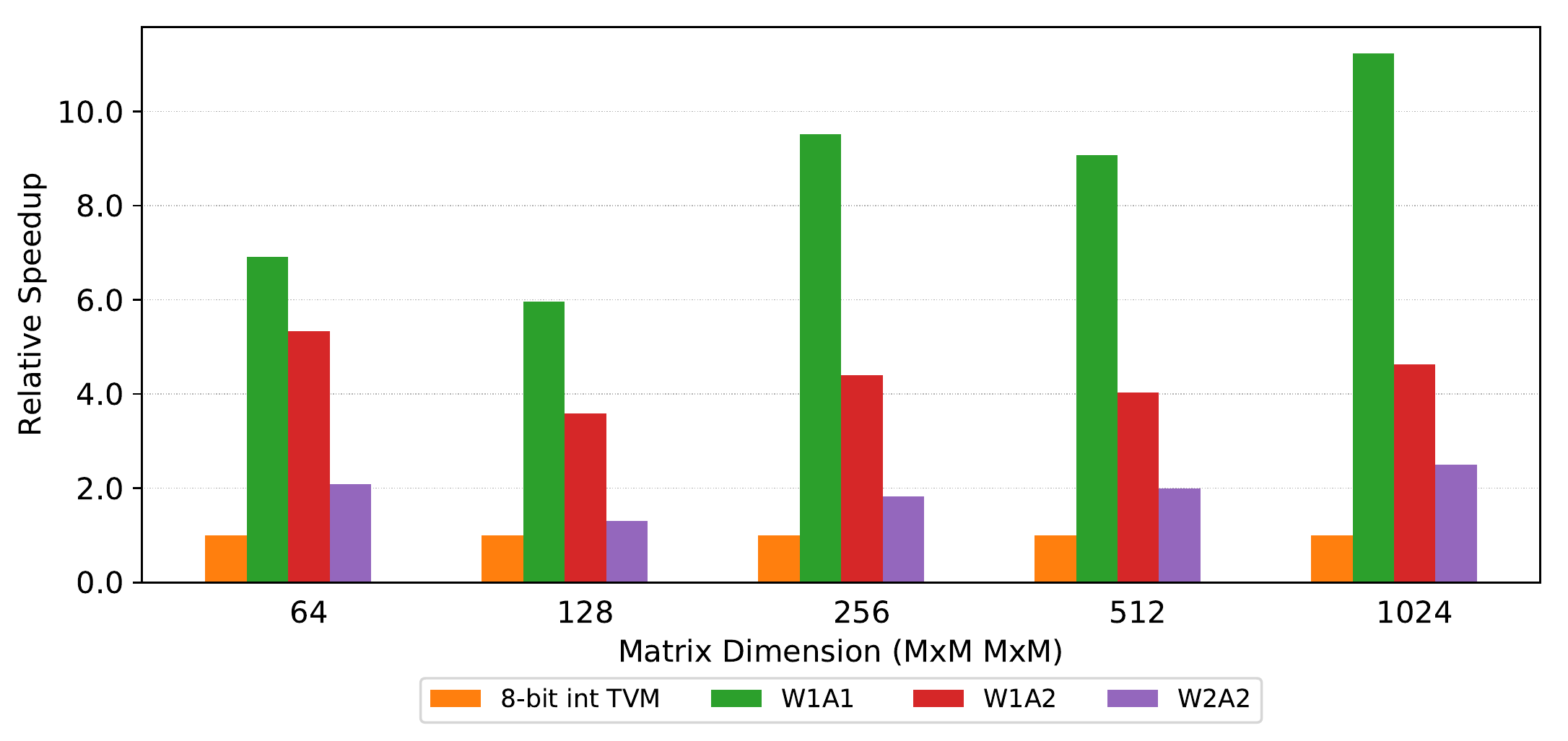}
\caption{Quantized matrix multiply speedup over an 8-bit integer baseline for the Raspberry Pi. The workload sweeps over multiplication of square matrices. }
\label{fig:matmul}
\end{figure}

We benchmark the performance of low precision matrix multiplication, which makeup the computation of dense fully connected layers, against an optimized 8 bit integer implementation in TVM, for three various precisions of W1A1, W1A2, and W2A2. Note the same schedule is used for all precisions, though tiling parameters have been tuned separately. 
To prevent saturation of results the outputs of both the 8-bit baseline and quantized operators stored a larger storage type, with the 8-bit baseline accumulating in 32-bit integers, and the quantized results accumulating in 16-bit integers.

In Figure \ref{fig:matmul}, we plot speedup relative to the 8-bit baseline. As expected, the W1A1 preforms significantly better and is up to 11x faster than the 8-bit baseline. The performance dropping steadily for higher quantization levels, with max speedups of 6.3x  and 3.1x for W1A2 and W2A2, scaling roughly with the product of bitwidths. 
\begin{table}[t]
  \begin{footnotesize}
	\begin{tabular}{ccccc}
	\hline
	Name & Operator & $H, W$ & $IC, OC$ & $K, S$ \\
	\hline
	2  & conv2d & 56, 56 & 64,64   & 3, 1 \\
	3  & conv2d & 56, 56 & 64,64   & 1, 1 \\
	4  & conv2d & 56, 56 & 64,128  & 3, 2 \\
	5  & conv2d & 56, 56 & 64,128  & 1, 2 \\
	6  & conv2d & 28, 28 & 128,128 & 3, 1 \\
	7  & conv2d & 28, 28 & 128,256 & 3, 2 \\
	8  & conv2d & 28, 28 & 128,256 & 1, 2 \\
	9  & conv2d & 14, 14 & 256,256 & 3, 1 \\
	10 & conv2d & 14, 14 & 256,512 & 3, 2 \\
	11 & conv2d & 14, 14 & 256,512 & 1, 2 \\
	12 & conv2d &  7,  7 & 512,512 & 3, 1 \\
	\hline
	\end{tabular}
  \end{footnotesize}
	\centering
	\caption{\small{Configurations of 2D-convolution operators in ResNet-18. Layer 1 is omitted as input channel depth is too small to allow efficient packing.
	H/W for height and width, IC for input channels, OC for output channels,
	K for kernel size, S for stride size.}}
    \label{tbl:all-op}
\end{table}

\begin{figure}
\centering
\includegraphics[width=0.5\textwidth]{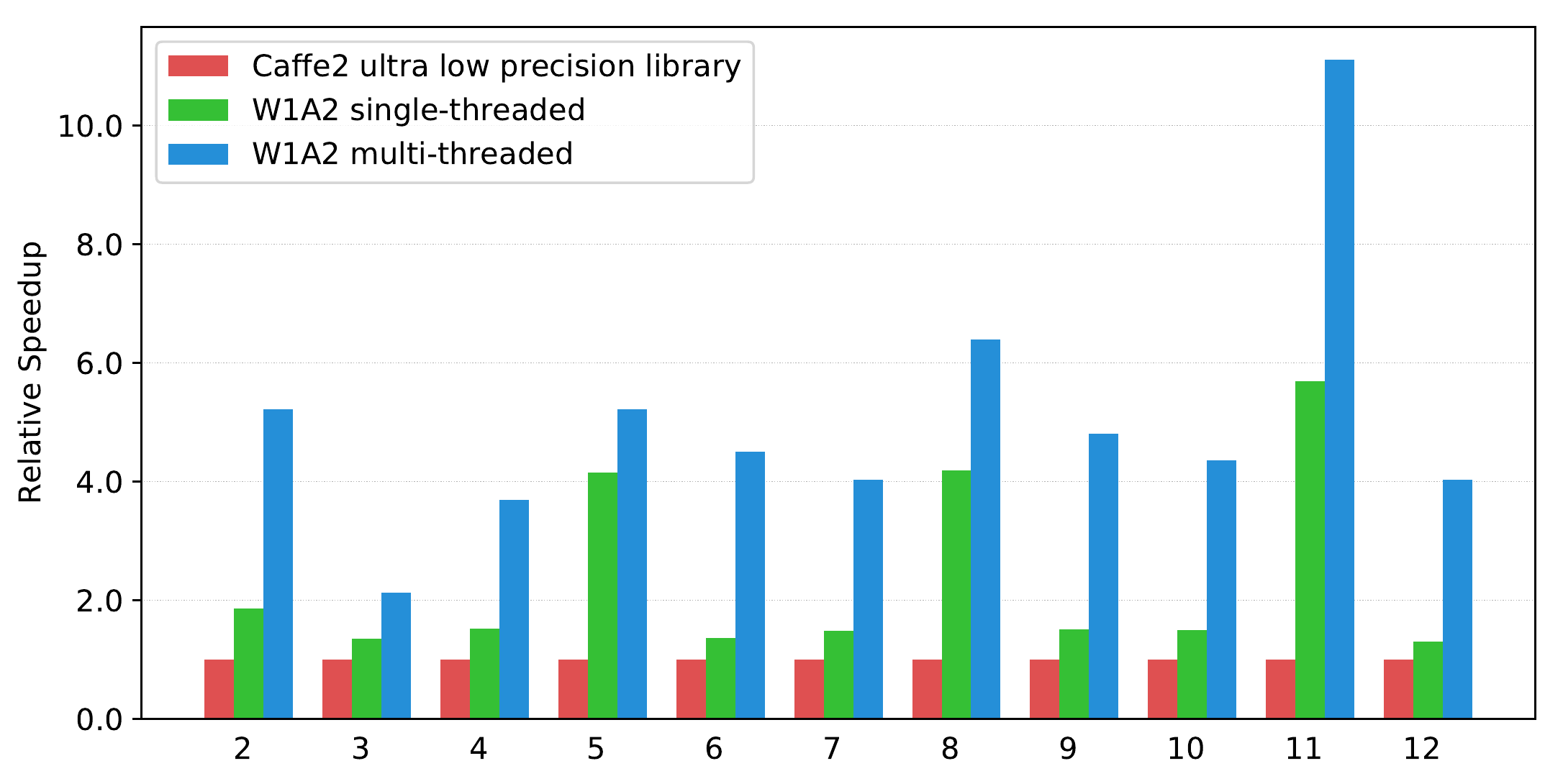}
\caption{Speedup of W1A2 convolution against a baseline Caffe2's hand optimized ultra low precision library for the Raspberry Pi. The baseline implementation is single threaded so for fair comparison we show a single threaded implementation and a multi-threaded implementation to show our maximum speedup.
}
\label{fig:fbconv}
\end{figure}

\begin{figure}
\centering
\includegraphics[width=0.45\textwidth]{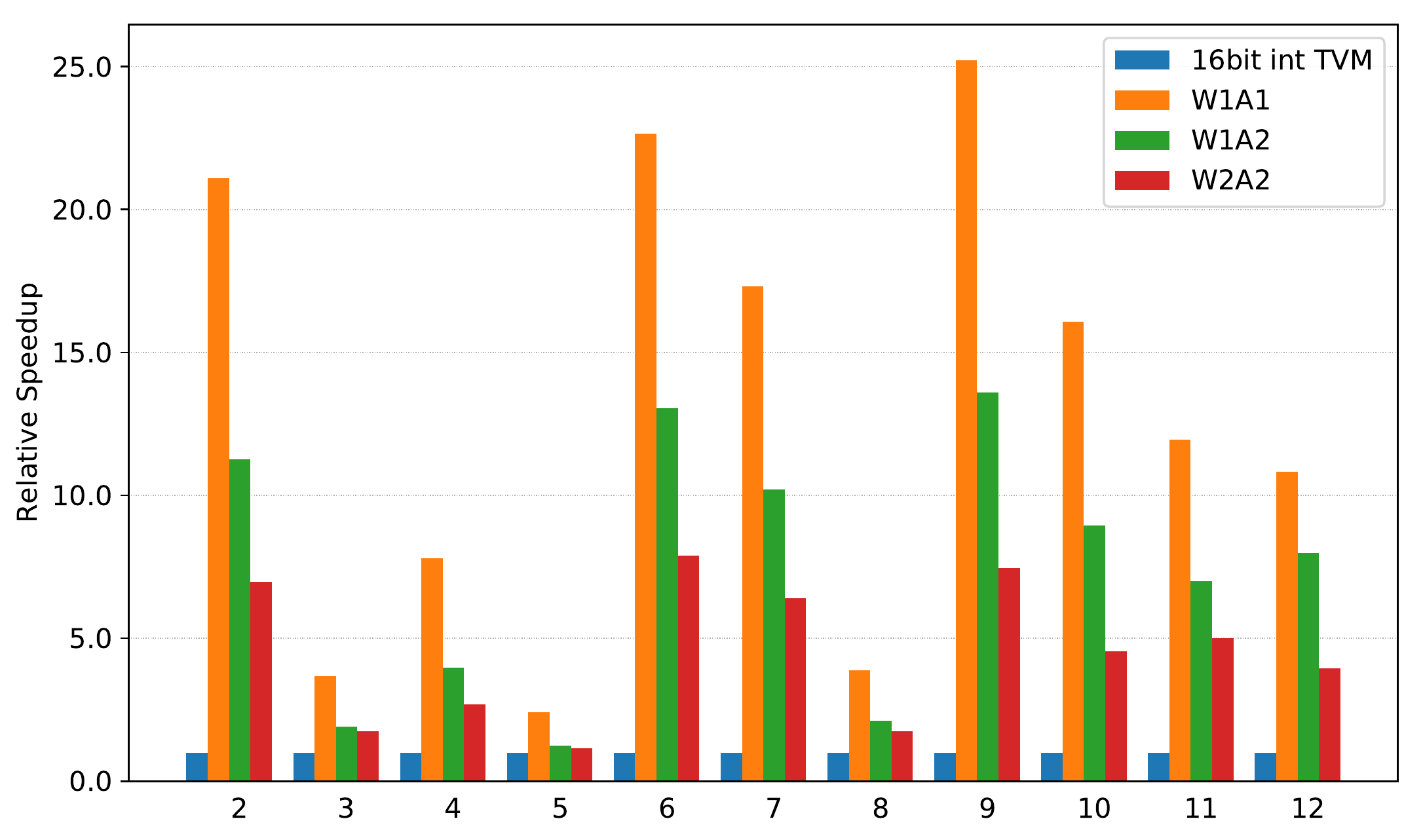}
\caption{Low precision convolution speedups on the Raspberry Pi, for various weight and activation precisions. The baseline is an optimized 16-bit integer implementation, and the workload is ResNet17, layers 2-12 (see Table \ref{tbl:all-op}).}
\label{fig:layer_perf}
\end{figure}

\begin{figure}
\centering
\includegraphics[width=0.5\textwidth]{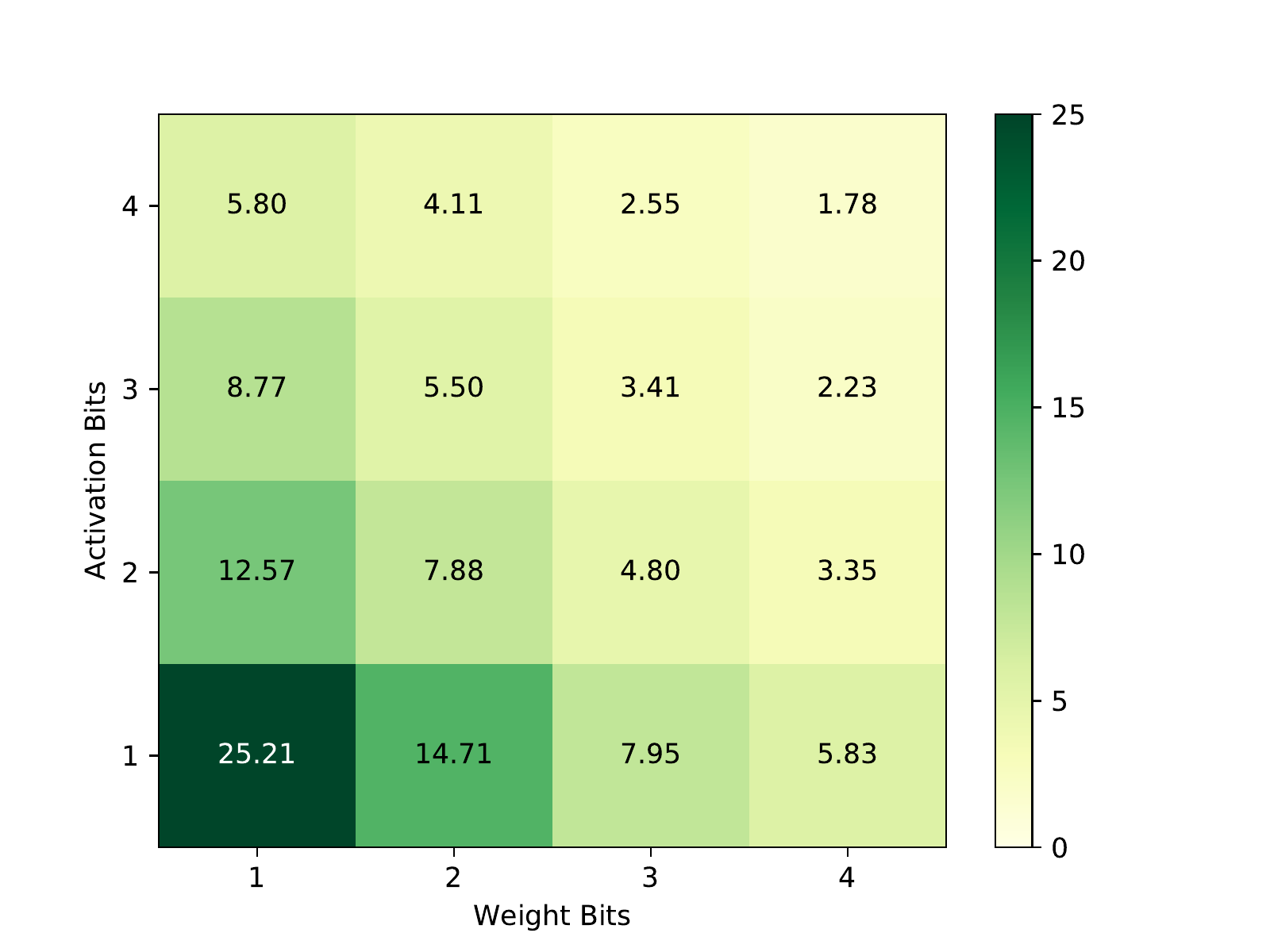}
\caption{Relative speedups of for Layer 9 of ResNet18 for combinations of a different data and weight bitwidths. Baseline is optimized 16-bit integer TVM baseline.}
\label{fig:rasp_limi}
\end{figure}

\begin{figure}
\centering
\includegraphics[width=0.5\textwidth]{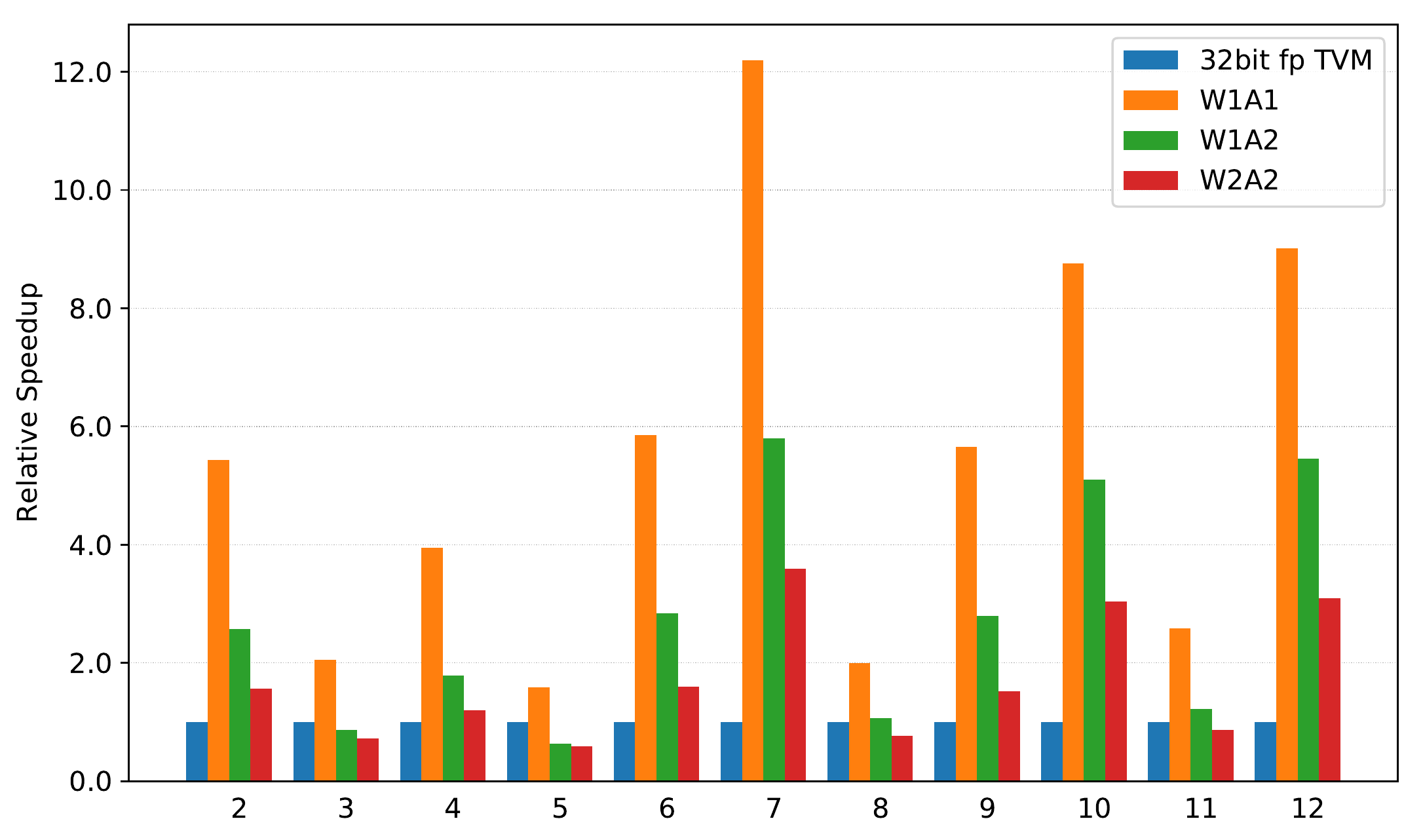}
\caption{Quantized convolution speedups for an x86 i7-4790K, for various weight and activation precisions. The baseline is an optimized 32-bit floating point, and the workload is ResNet17, layers 2-12 (see Table \ref{tbl:all-op}).}
\label{fig:x86}
\end{figure}

\subsubsection{Convolutions}
Convolutions makeup the bulk of image recognition neural networks. In this section we benchmark our low precision operators on layers 2-12 of ResNet18 (see Table \ref{tbl:all-op} for information). We omit the first layer, as most low precision networks preform the first layer in full precision. In Figure \ref{fig:layer_perf}, we plot the speedup of three low precision convolutions (W1A1, W1A2 and W2A2), against an optimized 16-bit integer baseline\cite{jiang2018efficient}, that experiences negligible accuracy degradation compared to full precision floating point. Our results show significant speedups over the baseline reaching 25x, 13x, and 8x speedups for W1A1, W1A2, and W2A2 respectively. 
It should be noted layers 3, 5, and 8 show poor speedups. However, these layers preform an order of magnitude fewer operations than other layers, indicating bitserial computation does not scale well for small problems that can't amortize the bit-packing costs.

Additionally, we compare our low precision convolutions against the current state of the art implementation, Caffe2's ultra low precision library\cite{tulloch2017high} that was written specifically for ARM v7/v8 architecture and in lines calls to ARM NEON intrinsics 
We verified their results that the inner-most loop of computation reaches about 70\% peak theoretical performance for binary operations, assuming a vectorized instruction can be issued every cycle. 

In Figure \ref{fig:fbconv}, we plot the speedup of our low precision implementations against theirs. The Caffe2 library currently only supports W1A2 operations and is single threaded, therefore we show a single threaded implementation of our code for fair comparison a multi-threaded implementation to highlight the maximum speedup we can achieve using parallelism.

Our single threaded implementation preforms roughly 2.3x better than the baseline. It should be noted that we preform significantly better on layers 5, 6, and 11, which are 1x1 convolutions with a stride of 2 that the baseline did not optimize well. Our implementation preforms 1.6x better when we omit these layers.
Since we modeled our tensorized inner-loop after their code, both implementation emit roughly the same assembly code for the inner-most loop. We attribute our single-threaded speedup to selecting better tiling parameters by using TVM's autotuning infrastructure, and the use of efficient in-place convolution that results in less memory duplication than the convolution lowering strategy the baseline implemented. 

\subsubsection{Bitserial Limit Study}
We preformed a low precision limit study to analyze how performance scales with increasing precision of weights and activations. In Figure \ref{fig:rasp_limi} we report speedup for all combinations of one to four bit activations and weight convolutions against the 16-bit integer baseline from section 6.2.2. We preform this study on Layer 9 of ResNet18 as it responded best to quantization.

Our results confirmed that computation scales roughly with the product of bitwidths, as the W1A1 convolution preforms 14.4X faster than W4A4. 
We also see the value of memory reuse as W1A4 and W2A2 have the same computational complexity, but the W2A2 preforming 1.3x better as we experience better data reuse among the bit planes. 



\subsection{X86}
We benchmark our low precision operators on layers 2-12 of ResNet18 (see Table \ref{tbl:all-op} for information) on X86 as well. 
Since we have not implemented any X86 specific schedules these results are for the generic CPU schedule. We compare our results against an optimized 32-bit floating point baseline implemented in TVM. These schedules are optimized to take advantage of X86's AVX2 vectorized operations and were contributed by engineers at Amazon. We do not compare against any hand optimized bitserial kernels as we currently do not know of any, perhaps due to X86 lack of a vectorized popcount instruction.

In Figure \ref{fig:x86} we plot the speed up of three low precision convolutions (W1A1, W1A2 and W2A2), against the full precision baseline. 
Our generic CPU schedule preformed moderately well achieving a average speedups of  
5.38x, 2.73x and 1.69x on W1A1, W1A2, and W2A2 respectively. Similar to the Raspberry Pi, we see a wide range of speedups across the layers, with peak speedups twice the average speedup, indicating some layers respond better to low precision than others. 


\section{Conclusion}
To conclude we detailed the modifications we made to TVM to implement low precision DNN operators for CPUs, emphasizing support for arbitrary low precision quantized neural networks. We provide a library of flexible compute operators such as \emph{bit-packing} and quantized bitserial convolutions and matrix multiply, that users can re-use or write develop custom schedules for.

We then used these operators to preform an extended case study on optimizing low precision bitserial convolutions for the Raspberry Pi, and showed that by using a custom schedule we could out-preform a hand optimized 1-bit weight 2-bit activation convolution kernel by 2.3x on the layers of ResNet. Furthermore, our 1-bit weight 2-bit activation end-to-end Raspberry Pi inference achieves a speedup of 3.3x over a full precision implementation. 

\section{Acknowledgements}
We would like to thank Yaman Umuroglu for his help guiding us, and Andrew Tulloch for letting us use his microkernel and providing feedback. This work was supported in part by a Google PhD Fellowship for Tianqi Chen,
ONR award \#N00014-16-1-2795, NSF under grants  CCF-1518703, CNS-1614717, and CCF-1723352, and gifts from Intel (under the CAPA program), Oracle, Huawei and anonymous sources.


\end{document}